\documentclass[10pt, conference]{ieeeconf}
\IEEEoverridecommandlockouts
\usepackage{cite}
\usepackage{amsmath,amssymb,amsfonts}
\usepackage{graphicx}
\usepackage{textcomp}
\usepackage{xcolor}
\usepackage[ruled,vlined, linesnumbered]{algorithm2e}
\usepackage{stackengine}  
\usepackage{booktabs}
\usepackage[skip=10pt, font=footnotesize]{caption}
\usepackage{siunitx}
\usepackage{hyperref}  
\hypersetup{
    colorlinks,
    linkcolor={red!75!black},
    citecolor={blue!75!black},
    urlcolor={blue!80!black}
}

\def\BibTeX{{\rm B\kern-.05em{\sc i\kern-.025em b}\kern-.08em
    T\kern-.1667em\lower.7ex\hbox{E}\kern-.125emX}}

\newcommand{\y}{
\mathbf{y}
}

\newcommand{\K}{
\mathbf{K}
}

\newcommand{\I}{
\mathbf{I}
}

\newcommand{\Rn}{
\mathbb{R}
}

\newcommand{\R}{
\mathbf{R}
}

\newcommand{\red}[1]{\textcolor{red}{#1}}

\DeclareMathOperator*{\argmin}{arg\,min} 

\newlength\mylen
\newcommand\myinput[1]{%
  \settowidth\mylen{\KwIn{}}%
  \setlength\hangindent{\mylen}%
  \hspace*{\mylen}#1\\}

\let\oldnl\nl
\newcommand{\nonl}{\renewcommand{\nl}{\let\nl\oldnl}}

\begin{document}

\title{Online Parameter Estimation for Safety-Critical Systems with Gaussian Processes}

\author{Mouhyemen Khan and Abhijit Chatterjee
\thanks{Mouhyemen Khan and Abhijit Chatterjee are with the School of Electrical and Computer Engineering, Georgia Institute of Technology, Atlanta, GA 30332, USA: {\tt\small \{mouhyemen.khan, abhijit.chatterjee\}@gatech.edu }
}
}

\maketitle
\begin{abstract}
Parameter estimation is crucial for modeling, tracking, and control of complex dynamical systems. However, parameter uncertainties can compromise system performance under a controller relying on nominal parameter values.
Typically, parameters are estimated using numerical regression approaches framed as inverse problems. However, they suffer from non-uniqueness due to existence of multiple local optima, reliance on gradients, numerous experimental data, or stability issues.
Addressing these drawbacks, we present a Bayesian optimization framework based on Gaussian processes (GPs) for online parameter estimation.
It uses an efficient search strategy over a response surface in the parameter space for finding the global optima with minimal function evaluations. The response surface is modeled as correlated surrogates using GPs on noisy data. The GP posterior predictive variance is exploited for smart adaptive sampling. This balances the exploration versus exploitation trade-off which is key in reaching the global optima under limited budget.
We demonstrate our technique on an actuated planar pendulum and safety-critical quadrotor in simulation with changing parameters. We also benchmark our results against solvers using interior point method and sequential quadratic program. By reconfiguring the controller with new optimized parameters iteratively, we drastically improve trajectory tracking of the system versus the nominal case and other solvers.

\end{abstract}

\section{Introduction}
Parameter estimation in complex dynamical systems is a challenging task. In the autonomous sector, accurate parameter estimation is crucial for reliable operation and control in many safety-critical applications, e.g., search and rescue, transport delivery \cite{autonomous_applications1}, \cite{autonomous_applications2}. Model uncertainties and inaccuracies lead to undesirable behavior in such systems with the potential of endangering health and lives. The problem is further compounded due to the evolving nature of dynamical systems. 
Regressor based methods, framed as inverse problems, are often used for estimating unknown parameters \cite{parameter_inverse_aster2018}. However, they require a lot of data and are typically limited only to offline settings.
Classical approaches in control theory typically use a model reference and estimate unknown parameters online using adaptation laws as ordinary differential equations \cite{adaptive_control_aastrom2013}. However, parameter drift is a key bottleneck in such techniques. 
In this paper, we propose a novel method of estimating parameters online based on a Bayesian optimization framework using Gaussian processes (GPs).


Parameter estimation is ubiquitous in science. A popular approach for estimating unknown parameters relies on inverse modeling \cite{parameter_inverse_aster2018}. Generally, in these methods a fitting optimization is performed based on numerical data for estimating unknown parameters. Inverse modeling techniques were used to estimate soil parameters as a multiobjective optimization problem \cite{param_soil_mertens2006} and for parameters determining flexible hydraulic functions \cite{param_hydraulic_durner1999}. However, these methods require a lot of data (which can be expensive), result in poor fitting without sufficient data, do not guarantee any global optima, or suffer from numerical instability. Moreover, techniques are limited to the offline setting which hinders their applicability for safety-critical systems.

Classical methods in control theory estimate parameters online using a reference model. With a nominal model of the plant running in the background, either controller gains can be updated (direct method) or plant parameters can be estimated first followed by updating the control laws (indirect method) \cite{adaptive_control_aastrom2013}. A composite approach combines both the methods giving better plant and controller estimates \cite{adaptive_quadrotor_dydek2012}. A problem arises when adaptation gains need to be very fast and high to deal with uncertainties. This leads to high frequency oscillations in the system amplifying noise or disturbances. $\mathcal{L}1$ adaptive control deals with fast adaptation by decoupling robustness and performance using a low-pass filter \cite{l1_adaptive_hovakimyan2010}. A key challenge well known in adaptive control is parameter drift during estimation \cite{adaptive_control_failures_anderson2005}. To address this challenge, persistency of excitation (PE) is required. Dynamical regressor extension and mixing was proposed for estimating parameters with convergence without requiring PE \cite{param_dynamic_regressor_aranovskiy2016}. However, this has many tunable degrees of freedom for ensuring convergence under strict conditions, such as, non-square integrable determinant (for linear regressors) and monotonicity (for nonlinear regressors).



With recent advances in machine learning, optimization algorithms have been developed that require no manual tuning. Bayesian optimization (BO) finds the global optima for a given cost function with minimal evaluations without making any gradient assumptions \cite{tutorial_bo_brochu2010}. It models the underlying cost function as a GP, and exploits its posterior variance for informative samples to reach the global optima. Another major advantage is that it explicitly models noise, which is pervasive in any practical system, without compromising the ability to reach the global optima. BO is most popularly used for hyperparameter tuning of learning algorithms \cite{practical_bo_ml_snoek2012}. BO also found its applicability in 
dynamical settings for
gait optimization of legged robots \cite{bo_gait_lizotte2007}, and controller optimization in snake-robot \cite{bo_snake_matthew2011} and quadrotor \cite{bo_quadrotor_berkenkamp2016} with very few evaluations. However, only controller gains were optimized instead of identifying unknown parameters which may contain important diagnostic information. For instance, estimating an unexpected change in mass allows not only controller modification but also necessary changes in path planning if desired, versus simply optimizing the controller gain while being agnostic to the root cause of the change. As far as we know, BO was used for estimating parameters only in arterial haemodynamic setting \cite{bo_haemodynamics_paris2016}; but restricted to offline case and systems which are not safety-critical.

\begin{figure*}[!t]
\centering
\includegraphics[width=1\linewidth]{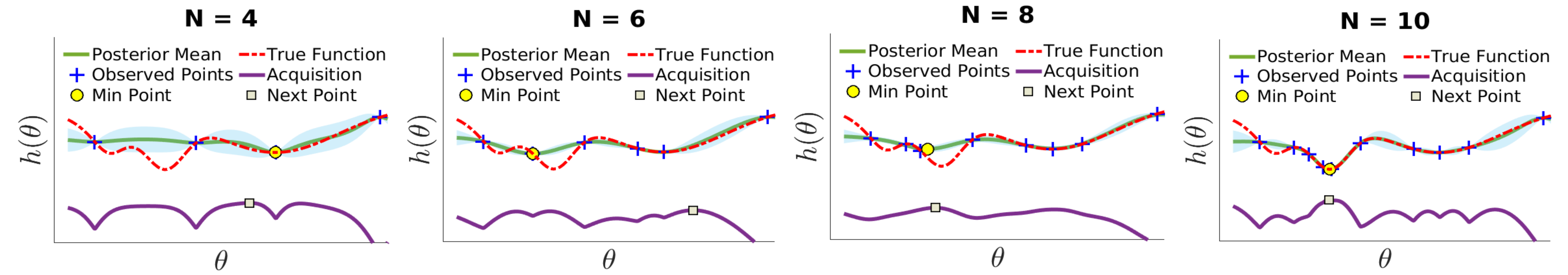}
\vspace{-0.7cm}
\caption{1D illustration of BO using GPs and EI acquisition function. The objective is to find the global minima of the unknown function (red-dashed). The top plot captures the posterior mean (green) and $2\sigma$ uncertainty (cyan-envelope) with observed training points (blue-plus). The bottom plot is EI function (violet) and proposed next sample (grey-square) balancing exploration-exploitation using EI. The global optima (yellow-circle) is reached under 10 evaluations.}
\label{fig:bo_sample_illustration}
\vspace{-0.5cm}
\end{figure*}
Our \textbf{key contributions} are the following. First, we perform parameter estimation online for safety-critical systems using GPs. The proposed paradigm leverages Bayesian statistics for training correlated surrogates of the underlying cost function while modeling noise to reach the global optima. To the best of our knowledge, BO was not used for parameter estimation online for safety-critical systems. Second, we model noise while reaching the global optima with very few evaluations without making gradient approximations. This addresses realistic scenarios since noise is pervasive and computing the gradient is always not possible. Third, the parameters estimated from the optimization routine are iteratively used to reconfigure the controller for a dynamically evolving system. This eliminates parameter drifting, since convergence is achieved iteratively to the global optima, while not having to invoke any PE condition.

The outline of the paper is as follows. Section \ref{sec:background} presents important preliminaries for GPs and BO. Section \ref{sec:param_est_gp} describes parameter estimation using GPs and controller reconfiguration. 
Simulation results are provided in Section \ref{sec:simulation}, followed by conclusions in Section \ref{sec:conclusion}.

\section{Background Preliminaries}\label{sec:background}
In this section, we provide important preliminaries surrounding GP regression and BO that is used throughout the paper.

\subsection{Gaussian Process Regression (GPR)}\label{subsec:gp}
We are interested in learning an underlying latent function $h(\theta)$, for which we assume noisy observations given by, $\varepsilon_i = h(\theta_i) + \omega_i$, where $\omega_i \sim \mathcal{N}(0, \sigma_{\omega}^2)$. Given a set of $n$ training points, with input vectors $\theta \in \mathbb{R}^d$, and scalar noisy observations $\varepsilon \in \mathbb{R}$, we compose the dataset:  $\mathcal{D}_n = \{ \mathbf{\Theta}_n, \mathbf{\mathcal{E}}_n \}$, where $\mathbf{\Theta}_n = \{\theta_i\}_{i=1}^{n}$ and $\mathbf{\mathcal{E}}_n = \{\varepsilon_i\}_{i=1}^n$. A GP places a distribution on the function $h(\theta)$, treating it as random variables associated with different values of $\theta$, any finite number of which produces a consistent joint Gaussian distribution \cite{Rasmussen2003_gpml}. 

A GP can be fully specified by its mean function $\mu(\theta)$ and covariance function $k(\theta,\theta')$. The latter is also called the kernel, measuring similarity between any two inputs $\theta, \theta'$. GPs can be used to predict the function value, $h(\theta_*)$, for an arbitrary query point $\theta_q$, by conditioning on previous observations. The posterior predictive mean $\mu$ and variance $\sigma$ are then given by \cite{Rasmussen2003_gpml}:
\begin{align}
\mu(\theta_q) &= k^{\top}_{n*} \big( \K_n + \sigma_{\omega}^2 \I_n \big)^{-1} \y_n \label{eq:gp_mean} \\
\sigma(\theta_q)^2 &= k (\theta_q, \theta_q) - k^{\top}_{n*} \big( \K_n + \sigma_{\omega}^2 \I_n \big)^{-1} k_{n*} \label{eq:gp_var},
\end{align}
where $k_{n*} = \big[k(\theta_1, \theta_q), \ldots , k(\theta_n, \theta_q) \big]^{\top}$ is the covariance between the input points in $\mathbf{\Theta}_n$ and query point $\theta_q$, $\K_n \in \mathbb{R}^{n \times n}$ has entries $\big[ \K_n \big]_{(i,j)} = k(\theta_i, \theta_j), \ i, j \in \{1, \ldots, n\}$, is the covariance matrix between pairs of input points, and $\I_n \in \mathbb{R}^{n \times n}$ is the identity matrix. The hyperparameters of a GP depend on the kernel choice and can be problem-dependent. We refer the reader to \cite{Rasmussen2003_gpml} for a review of different kernels. The values of the hyperparameters that best suit the particular dataset can be derived by maximizing the log marginal likelihood using quasi-Newton based methods , e.g., L-BFGS \cite{Rasmussen2003_gpml}.

\subsection{Bayesian Optimization (BO)}\label{subsec:bo}
Now consider another unknown function $g$, whose global optima we are interested in. The objective of BO is to find the global optima of such an unknown function. The function $g$ could represent the cost function or the performance measure of a system which could be expensive to evaluate. Since $g$ is unknown, finding its global optima is an impossible task. Hence, several assumptions are in order. 

The objective function $g$ is typically considered to be a black box function; we may not have its analytical expression or derivatives. Bounds are placed on the possible search space to find the optima. Evaluation of the function is restricted to querying at a sample $\theta$ and observing a (noisy) estimate. With these assumptions in mind, a surrogate predictor $h(\theta)$ replaces the expensive function $g$ (see \ref{subsec:gp}) . GPs are popular surrogate predictors among several candidates, such as neural networks or radial basis functions, due to its Bayesian non-parametric nature. GPs also offer a flexible prior over function, robust analytical tractability, and high probabilistic guarantees \cite{Rasmussen2003_gpml}.

In general, BO models the unknown function $g$ using the GP posterior mean given by (\ref{eq:gp_mean}). GP's probabilistic structure enables an efficient search strategy for the next sampling point by exploiting its predictive variance (\ref{eq:gp_var}). This gives rise to what is called an acquisition function. By efficiently searching over the space using the acquisition function, exploration-exploitation trade-off is addressed to find the global optima. There are several acquisition functions to choose from, e.g., probability of improvement (PI), expected improvement (EI), lower-confidence bound (LCB). We refer the reader to \cite{tutorial_bo_brochu2010} for detailed analysis of different acquisition functions and their tradeoffs. Figure \ref{fig:bo_sample_illustration} demonstrates the sequential nature of BO for finding the global minima of an unknown function using EI acquisition function.

\section{Parameter Estimation using Gaussian Process}\label{sec:param_est_gp}
Here, we describe our proposed strategy for estimating unknown parameters for a dynamical system using GPs and BO.

\subsection{Bayesian Optimization Formulation}\label{subsec:bo_formulation}
We consider a nonlinear, continuous-time system, 
\begin{align}\label{eq:nonlinear_system}
\dot{x} &= f(x(t), u(t) ; \theta_{p} (t) ),
\end{align}
where $x(t) \in \mathcal{X} \subset \R^n$ is the state, $u(t) \in \mathcal{U} \subset \R^m$ is the control input, and $\theta_p(t) \in \Theta \subset \R^d$ are the system parameters at time $t \in \Rn$. We assume a nominal control policy $u_{nom}(t; K, \theta_{n})$ exists, where $K$ are the controller gains and $\theta_n \in [\theta_n^{min}, \theta_n^{max}]$ are the nominal system parameters at time $t$. $u_{nom}$ is designed to drive $f(x,u; \theta_p)$ to the zero-equilibrium stable point under nominal conditions. Ideally, $\theta_n$ is designed to be nominally close to $\theta_p$. However, $\theta_p$ can change, thereby degrading the controller performance since it depends on $\theta_n$. Hence, the controller's $\theta_n$ must be updated to better approximate $\theta_p$.

The goal is to estimate the time-varying $\theta_p(t)$
and reconfigure the nominal controller's $\theta_n$ with the new estimate \textit{online}. To learn the unknown estimate $\theta_p(t)$, we construct a model of the dynamical system $\hat{f}(\theta^*) = f(x,u ; \theta^*)$ to match a target response $y^*$. This translates to the following optimization problem,
\begin{align}\label{eq:opt_problem}
\argmin_{\theta^* \in \R^d} g \hspace{0.125cm} \text{where } g \approx h(\theta) := \overbrace{	
\lVert 
\underbrace{y^*}_{\text{\footnotesize target}} 
- \underbrace{\hat{f}(\theta^*)}_{\text{\footnotesize model}} 
\rVert 
+ \underbrace{\mathcal{N}(0,\sigma_{\omega}^2)}_{\text{\footnotesize noise}}
}^{\text{\footnotesize Modeled using GP surrogate}},
\end{align}
in some suitable norm, where $\theta^* \in [\theta^*_{min}, \theta^*_{max}]$ are the estimated parameters. Here $g$ is an unknown and complex nonlinear map representing the error response surface. To alleviate these complexities, we use GPs as surrogate predictors to construct $h(\theta)$ approximating the true unknown surface $g$. We incorporate knowledge of the dynamics and noise as seen in (\ref{eq:opt_problem}) for more accurate posterior constructions of the response surface using GPs. 
Here, $y^*$ represents the noisy measurements of the system dynamics $[\dot{x} + \mathcal{N}(0,\sigma_{x}^2)]$. 




For a complex dynamical system, it is not feasible to perform numerous optimization steps requiring immense data. This is because the system is already deployed and reaching the feasible solution with minimal evaluations is crucial. 
Hence, the sampling strategy needs to be efficient and exploitative in its approach to reach the global optima for a time-varying dynamical system.

\subsection{Gaussian Process Modeling}\label{subsec:methodology}
Now that the minimization objective has been formulated, we next look at the design of the GP surrogate and acquisition function. The surrogate agent $h(\theta)$ is responsible for modeling the underlying unknown function $g$, whereas the acquisition function is responsible for guiding the sampling strategy at new query points.

Since GPs act as an intermediate agent to model $g$ in (\ref{eq:opt_problem}), a kernel function is required for describing GPs. In this work, we choose from the family of Mat\'ern kernels. Owing to its flexible parameterization and simplicity, this family of kernels is commonly employed in the literature \cite{Rasmussen2003_gpml}. An attractive property of the Mat\'ern kernel is in its ability to model a wide spectrum of responses ranging from infinitely differentiable functions to rough  Ornstein-Uhlenbeck paths. The Mat\'ern $5/2$ kernel is given by,
\begin{align}\label{eq:matern_kernel}
k(\theta_i, \theta_j) &= \sigma_f^2  \big( 1 + \sqrt{5}r + 5r^2  \big) \exp \big( -\sqrt{5}r \big) \\
r(\theta_i, \theta_j) &= \sqrt{ \big( \theta_i - \theta_j \big)^{\top} \mathbf{L}^{-2} \big( \theta_i - \theta_j \big) }, \notag
\end{align}
where $\theta_i$ and $\theta_j$ are any two samples, $\sigma_f$ is process variance, $r$ is the Euclidean distance between the samples weighted by $\mathbf{L}$ which is a diagonal matrix with elements, $l \in \Rn_+^{|\Theta|}$, corresponding to separate length scales for each $\theta$. 
Hence, the kernel used is the automatic relevance determination Mat\'ern $5/2$ kernel.

Next, the acquisition function is to be determined which is key in navigating the sampling strategy for finding the global minima $\theta^*$ in (\ref{eq:opt_problem}). The construction of the acquisition function uses the GP posterior prediction (\ref{eq:gp_mean}-\ref{eq:gp_var}) parameterized by a kernel of choice (\ref{eq:matern_kernel}). The acquisition function then informatively samples new query points in order to balance the exploration-exploitation divide, i.e., searching globally unexplored places and minimizing uncertainty versus exploiting a local search expecting the global optima to reside nearby. In this work, we focus on the expected improvement (EI) function which is one of the most commonly used acquisition functions addressing the exploration-exploitation mindset. The improvement function originally proposed in \cite{expected_improvement_mockus1994} is,
\begin{align*}
I(\theta) = \max \{ 0, h_{t+1} (\theta) - h(\theta^+) \},
\end{align*}
where $h_{t+1}(\theta)$ is the next evaluation, $h(\theta^+)$ is the best posterior mean from GP on past points or the best observation so far. Intuitively, this evaluates to a positive improvement when the prediction is higher than the current best value, else it is set to zero. By maximizing the improvement function, the next sample's location is then found,
\begin{align}\label{eq:ei_acq_func}
\theta^{\text{next}}= \argmin_\theta \mathbb{E} (I(\theta) | \mathcal{D}_t),
\end{align}
where $\mathcal{D}_t$ represents all the observations collected thus far. During exploration phase, the function would choose points where the surrogate variance is large. During exploitation phase, the function would choose points where the surrogate mean is large.

\subsection{Controller Reconfiguration}\label{subsec:controller_reconfiguration}
The primary objective is to estimate unknown parameters in an \textit{online fashion} and update the controller for a dynamical system. While BO reaches the global optima in few evaluations, which is very attractive for our interests, waiting for it to converge to the global optima overrules the online requirement. Hence, it is imperative to update the controller iteratively while the optimizer converges to the global optima. The nominal controller $u_{nom}(t; K, \theta_n)$ then takes on the form $u(t; K, \theta^*)$ to keep the system $f(x,u; \theta_p)$ operating under nominal conditions. 
Here, the nominal operating behavior can be quantified by bounding below an error threshold given by, $e(t) := x_{ref}(t) - x(t) \leq \tau_e$, where $x_{ref}(t)$ is a desired reference trajectory, and $x(t)$ are the states of the system. If the threshold is exceeded, then the optimizer would solve for finding the unknown estimates.


We construct the termination criteria of the optimizer in terms of number of evaluations, $N$. Due to the explorative-exploitative approach to reach the global optima, before $N$ evaluations are completed, it is possible to return parameter estimates that are critical to the system since the controller is reconfigured at every step. Therefore, this may allude one to posing the following question: \textit{As new estimates returned from the optimizer are used to reconfigure the controller, would the system remain safe?} 

This problem is reasonably alleviated given fast computations since each evaluation time will be faster (several $\si{\hertz}$) than the physical time constant of the system. Since the EI function samples for locations with the improvement criteria, the controller progressively moves to better parameter estimates. Hence, it does not use (potentially) unsafe parameters for a duration longer than the physical time constant of the system. Finally, after $N$ steps, the controller eventually commits to the best parameter estimates, rendering the system to a better condition than the nominal case.

\subsection{Algorithm Overview}\label{subsec:algorithm}
We summarize our proposed framework of finding unknown parameters online for safety-critical systems as shown in Algorithm \ref{algo:param_est_outline}. The algorithm is initialized with a domain for unknown parameters, reference trajectory for a time period, and error threshold with maximum iteration budget for initiating or terminating the optimizer.

The proposed framework's main actions happen from lines $6-20$. As seen from the algorithm, if the trajectory error does not exceed an error threshold, then the nominal controller designed on line $3$ suffices. If the threshold is exceeded, then the GP posterior is used to efficiently find the globally optimized unknown parameters iteratively (lines $10-18$). 

An initial dataset is constructed using random parameters and their responses (lines $7-9$).
Using the GP mean and variance (\ref{eq:gp_mean}-\ref{eq:gp_var}) in line $11$, the EI function (\ref{eq:ei_acq_func}) is used to query the next parameter estimate with the highest improvement (line $12$). While (\ref{eq:ei_acq_func}) is also an optimization problem, it only relies on the GP model and does not require any evaluations on the real dynamical system. This corresponds to cheap computations and can be optimized quickly. The new parameter estimate is then used to reconfigure the controller in line $13$. Noisy observations are embedded while observing the system states (lines $4$ and $14$). An interation budget of $N$ helps tune the number of function evaluations to reach the global optima. 


\begin{algorithm}[!b]
\caption{Online parameter estimation using GPs and BO}
\label{algo:param_est_outline}
	\SetAlgoLined
	\KwIn{Domain: $\Theta$ 			\hspace{1.00cm} GP Prior: $\mathcal{N} \big( 0, k(\theta_i, \theta_j) \big) $}
	\nonl \myinput{Reference: $x_{ref}$ 	\hspace{0.4cm} Error threshold: $\tau_e$}
	\nonl \myinput{Period: [$t_0, t_f$] \hspace{0.55cm} Iterations: $N$}
	\textit{Initialize:} $t = t_0$ \\
	\While{$t \leq t_f$} {
	Compute controller: $u_{nom}(t; K, \theta_n)$ \\
	Observe states: $x_{t}$ \\
	Compute error: $e_{t} \gets  x_{ref,t} - x_{t}$ \\
	 	\If{$e_t \geq \tau_e$}{
	 		Randomly select $\theta_{rand} \in \Theta $ \\
			Observe dynamics: $y_t \gets \dot{x} + \mathcal{N}(0,\sigma_x^2)$  \\
	 		Error response: $\varepsilon_{rand} \gets \lVert y_{t} - f(x,u; \theta_{rand}) \rVert$ \\
			Dataset: $\mathcal{D}_{t} \gets \{\theta_{rand}, \varepsilon_{rand}+ \sigma_{\omega}\} $ \\
			\For{$i = 1 \hdots N$} {
				Compute GP: $\mathcal{GP} = \mathcal{N}(0,k(\theta_i, \theta_j) ; \mathcal{D}_t)$ \\
				Evaluate EI: $\theta_t = \argmin_\theta \mathbb{E} (I(\theta) | \mathcal{D}_t) $\\
			  	Update controller: $u_t \gets u(t; K, \theta_t)$ \\
				$y_t \gets \dot{x} + \mathcal{N}(0,\sigma_x^2)$  \\
				$\varepsilon_{t} \gets \lVert y_{t} - f(x,u; \theta_t) \rVert + \sigma_{\omega}  $ \\
				Update GP: $ \mathcal{GP} \big( \mathcal{D}_{t} \gets \{\theta_t, \varepsilon_t\} \big)$  \\
				Update dynamics: $x_t \gets x_t + \dot{x} \delta t $\\
			}
 		}
	Update dynamics: $x_t \gets x_t + \dot{x} \delta t $\\
	}
\end{algorithm}

\section{Simulation Results}\label{sec:simulation}
Here, we aim to demonstrate the effectiveness of our method through two systems. In the first scenario, we study the case of an actuated planar pendulum with changing parameters, with the intent of providing a pedagogical example highlighting the main aspects of our work. In the second scenario, we take a more realistic example of a safety-critical quadrotor. The quadrotor is an ideal candidate for validating the efficacy of our method due to its inherently underactuated structure and highly nonlinear dynamics evolving in the tangent bundle to the Special Euclidean group. As the parameters change in the quadrotor, it easily deviates from the reference trajectory, or worse, performs erratically and becomes unstable. Hence, it is important to accurately estimate the unknown parameters as quickly as possible for the quadrotor to remain in stable flight operation.

\subsection{Actuated Planar Pendulum}\label{subsec:pendulum}
Consider a planar pendulum with friction that has a single degree of actuation. It has two states, $[\phi, \ \dot{\phi}]$ which are the angular position and velocity of the pendulum respectively. The dynamics is given by,
\begin{align}\label{eq:pendulum_dynamics}
\dot{x}_1 &= x_2 \\
\dot{x}_2 &= -\frac{g}{l} \sin(x_1) - \frac{b}{m}x_2 + \frac{u}{ml},
\end{align}
where $x = [x_1, \ x_2]^{\top} := [\phi, \ \dot{\phi}]^{\top}$ is the state of the system, $m$ is the pendulum mass, $l$ is the length of the rod, $b$ is the friction coefficient, and $u$ is the control input. Feedback linearization is used to design the nominal controller as shown below,
\vspace{-.1cm}
\begin{align}\label{eq:pendulum_controller}
u &= 	ml \Big( K_p e_1 + K_d e_2 \Big) +	ml \bigg( \frac{g}{l} \sin(x_1) + \frac{b}{m}x_2 \bigg),
\end{align}
where $e_i = x_{ref,i} - x_i$, $i = \{1, 2\}$, $K_p$ and $K_d$ are positive gains. 
To enable a meaningful visualization of the process, we confine ourselves to the study of a two-dimensional unknown parameter space. Let us consider that the pendulum's mass and length change at an unknown time. The cost function using knowledge of the dynamics and resulting in the GP surrogate's response surface is given by,

\begin{align*}
\min_{\theta^*} 
\Big\lVert 
	\underbrace{\dot{x}_2}_{\text{dynamics}} \ 
	- \ \underbrace{\frac{g}{l^*} \sin(x_1) - \frac{1}{m^*} \Big(bx_2 + \frac{u}{l^*} \Big)}_{\text{\footnotesize model}}
\Big\rVert 
	+ \underbrace{\mathcal{N}(0,\sigma_{\omega}^2)}_{\text{\footnotesize noise}}
\end{align*}
where $\dot{x}_2$ is the observed dynamics and $\theta^* = [m^*, l^*]$ are the design parameters.
The nominal mass and length are changed from $1.75 $ to $4.271 $ $\si\kilogram$ and $0.75$ to $0.981$ $\si\meter$ at $t = 3 $ $\si\second$. The reference is set as $x_{\text{des}} = [60^{\circ}, 0 \ \si\radian\si/\si\second]$. The domains set are $[0.1, 5.0]$ $\si\kilogram$ for mass and $[0.1, 3.0]$ $\si\meter$ for length, $N=30$, and $\tau_e = 0.01$. The controller in (\ref{eq:pendulum_controller}) is iteratively modified with new parameters, $u(x; K, \theta_n) \rightarrow u(x; K, \theta^*)$ at every time step until $N$ evaluations are reached.

\begin{figure}[!t]
\centering
\includegraphics[width=0.95\linewidth]{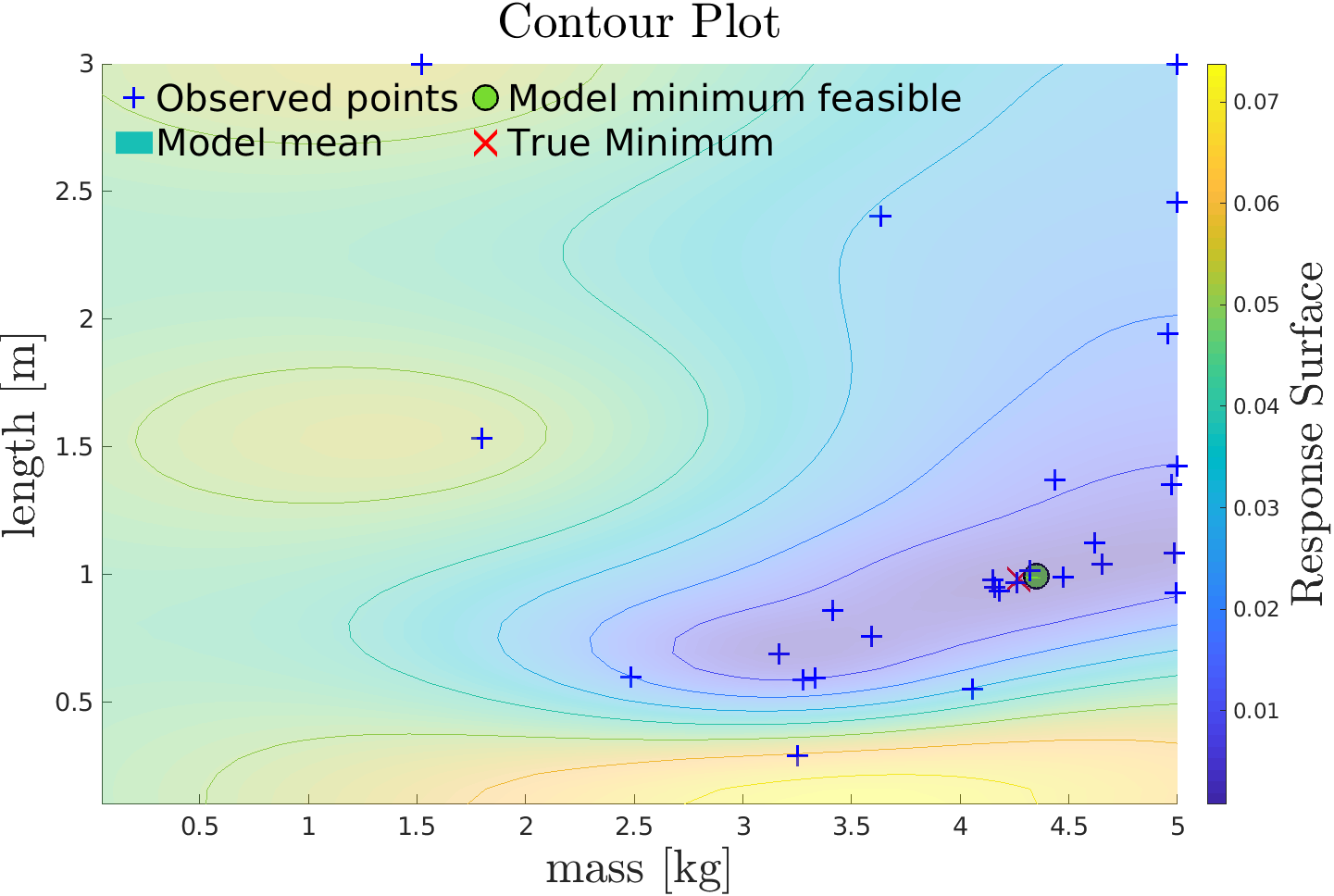}
\vspace{-0.25cm}
\caption{BO solves for the unknown mass and length under $N=30$ evaluations.}
\label{fig:pendulum_solution}
\end{figure}

\begin{figure}[!t]
\vspace{-.35cm}
\centering
\includegraphics[width=0.9\linewidth]{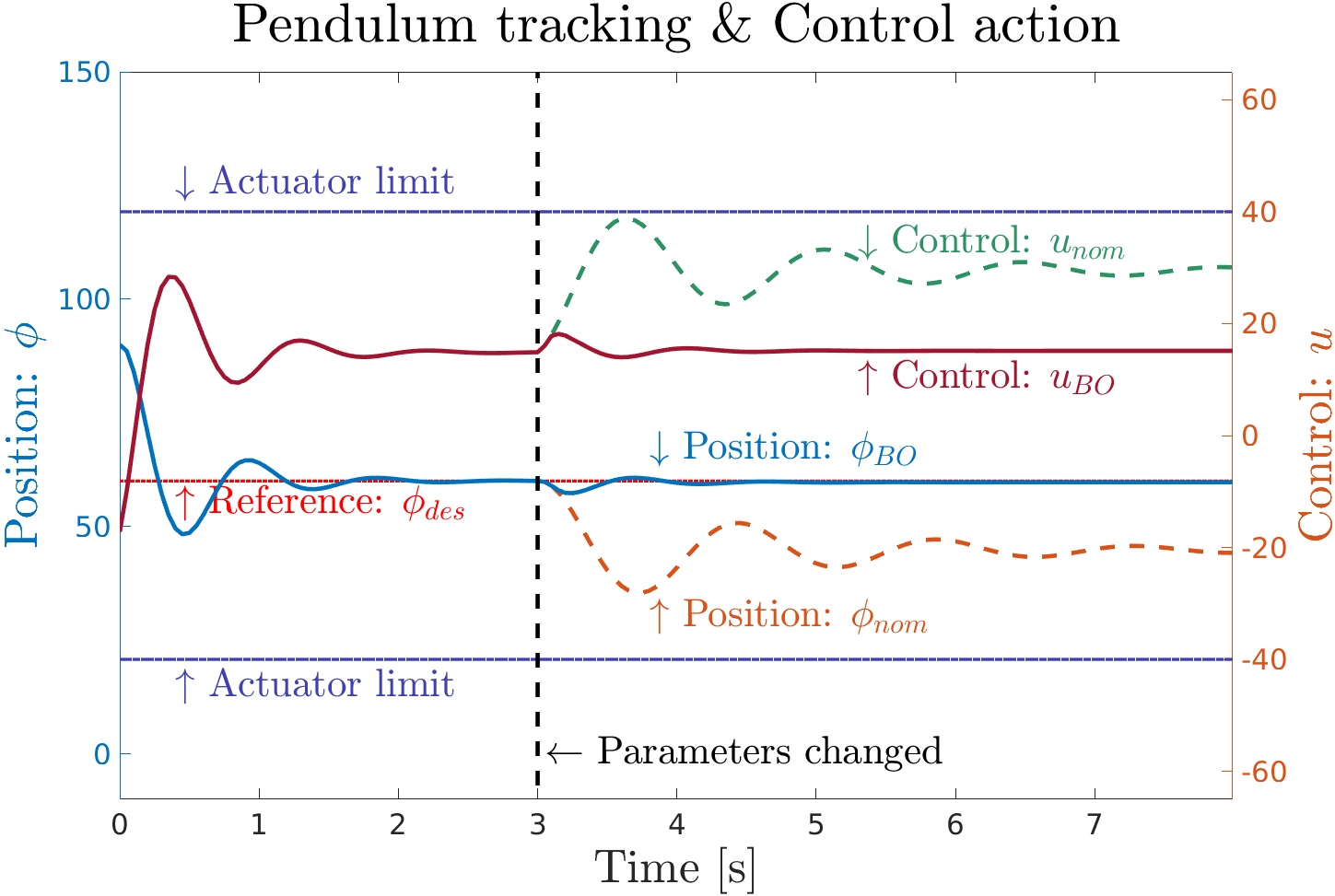}
\caption{The reconfigured controller ($u_{BO}$) tracks the reference position $\phi_{des}$ unlike the nominal controller ($u_{nom}$). $\phi_{BO}$ and $\phi_{nom}$ are the angular positions for system with estimated parameters and nominal system respectively.}
\label{fig:pendulum_tracking}
\end{figure}

As seen in Figure \ref{fig:pendulum_solution}, just under $30$ evaluations, the unknown parameters are estimated very close to the true minima. The deviation can be attributed due to modeling of noise, since the GP surrogates try to estimate the underlying true function, and is not a true representation. Combinations of masses and lengths are observed in an explorative manner to find the global estimate. Since the evaluations happen quickly, the controller progressively commits to better estimates and leads to a more stable tracking performance as seen in Figure \ref{fig:pendulum_tracking}. The controller is given parametric combinations far away from the true value for small time steps during the exploration phase. Despite this, the reconfigured controller tracks the reference due to progressively better reconfiguration unlike the nominal case.

\subsection{3D Quadrotor}\label{subsec:quadrotor}
Here, we do a more thorough analysis of our proposed solution for a highly nonlinear and complex safety-critical system. We first present the dynamics and design of the nominal controller for the quadrotor. We then validate our parameter estimation framework in a four-dimensional setting where the mass and inertia matrix of the quadrotor are varying with time. This is a particularly challenging setting since a quadrotor is a highly unstable system and is very susceptible to parametric changes. Hence, accurate online parameter estimation is imperative. The simulation was done in MATLAB 2019.

We consider the complete dynamics of a quadrotor evolving in a coordinate-free framework. This framework uses a geometric representation for its attitude given by a rotation matrix $\R$ in Special Orthogonal group, $SO(3) := \{ \R \in \Rn^{3 \times 3} | \R^{\top}\R = \I, \ \det(\R) = 1 \}$. 
The quadrotor has $4$ control inputs; thrust $F \in \Rn$ and moments $M \in \Rn^3$. The dynamical equations of motion are given by \cite{quadrotor_geometric_lee2010}:
\begin{align}
\dot{v}		&= g e_3 - \frac{F}{m} \R e_3 	\label{eq:lin_acc}\\
\dot{\Omega}	&= J^{-1} \big( M - (\Omega \times J \Omega) \big)	\label{eq:ang_acc}
\end{align}
where $v$ is quadrotor velocity in inertial frame, $m$ is mass of the quadrotor, $g$ is gravity, $J \in \Rn^{3 \times 3}$ is inertia matrix of the quadrotor, $e_3 = [0 \ 0 \ 1]^{\top}$, and $\Omega \in \Rn^3$ is body-frame angular velocity. For a complete mathematical treatment for the nominal controller, see \cite{quadrotor_geometric_lee2010}. Here, we simply present the equations for nominal $F$ and $M$:
\begin{align}\label{eq:quadrotor_controller}
F &= (-K_r e_r - K_v e_v + mge_3 + m\ddot{r}_d)^{\top} \cdot Re_3 	\notag \\
M &= - K_\R e_\R - K_{\Omega} e_{\Omega} + \Omega \times J\Omega 				\\ 
 	 & \ - J(\Omega^{\times} \R^{\top}\R_d \Omega_d - \R^{\top}\R_d \dot{\Omega}_d)  \notag,
\end{align}
where $K_{(\cdot)}$ are positive definite gains, $e_r = r - r_d$, $e_v = v - \dot{r}_d$, $e_{\R}		= \frac{1}{2}(\R_d^{\top}\R - \R^{\top}\R_d)^\vee$, and $e_{\Omega}	= \Omega - \R^{\top}\R_d\Omega_d$. The quadrotor's inertial position and velocity are $r$ and $v$. The desired position, velocity, orientation, and angular acceleration are $r_d, \dot{r}_d, \R_d,$ and $\dot{\Omega}_d$ respectively. $(\cdot)^{\times}$ is the skew-symmetric operator satisfying, $\forall a,b \in \Rn^3, a^{\times}b = a \times b$, and $(\cdot)^\vee$ is its inverse, i.e., $(a^{\times})^\vee = a$.

References are sinusoids where position reference is $[x_d, \ y_d, \ z_d]^{\top} = [4 \sin(0.8 t), 5 \sin(0.4 t), 2 \sin(0.4 t)]^{\top}$ and desired yaw is $\psi_d(t) = \text{atan2}(y_d,x_d)$, for $t \in [t_0, t_f]$. Nominal parameters are $m = 1.25$ $\si\kilogram$, $J = \text{diag}[1.1,  \ 1.1, \ 2.2]$ $\si\kilogram \si\meter^2$, gains $K_r = \text{diag}[5, 5, 5]$, $K_v = \text{diag}[0.5, 0.5, 2.0]$, $K_{\Omega} = \text{diag}[5, 10, 20]$, $K_{\R} = \text{diag}[30, 30, 30]$. The unknown time-varying mass and inertia are given by,
\begin{align}
\hat{m} &= 
\begin{cases} 
      \ 0.2\exp(-0.2t)\sin(1.5t) + m, 		\hspace{0.30cm}  	t \leq  3 		\\
      \hspace{1cm} 1.85,					\hspace{3.00cm} 	t \leq  6 		\\
      0.7\exp(-0.2t)\sin(1.5t) + m, 		\hspace{0.40cm}  	t \leq  9 		\\
      \hspace{1cm} 2.10, 					\hspace{3.00cm} 	t \leq 12 		\\
      \hspace{1.3cm} m	 					\hspace{3.20cm}    t \leq t_f
\end{cases}	\label{eq:time-varying_mass}
\\
\hat{J} &= 3.0 \ (J + \hat{m}r^2), 		\hspace{2.60cm} 	t_0 \leq t \leq t_f\label{eq:time-varying_inertia}
\end{align}
where $\hat{m}$ and $\hat{J}$ are the affected mass and inertia parameters respectively, $r = [0.2 \ \ 0.2 \ \ 0.2]^{\top}$ is the inertial axis offset, $t_0 = 0 \ \si\second$, and $t_f = 16 \ \si\second$. The goal is to estimate varying parameters for both slowly-varying and transient changes, since this demonstrates if parameter estimation is robust to different classes of changes. We assume inertial parameters form a diagonal matrix, hence the optimizer solves for an unknown inertial value along each principal axis. The response surface is constructed as follows,
\begin{align}\label{eq:quadrotor_opt}
\min_{\theta^*} 
\Big\lVert 
	\underbrace{\dot{x}(t)}_{\text{\small dynamics}}  \
	- \ \underbrace{f(x, u; \theta^* )}_{\text{\small model}}
\Big\rVert 
	+ \underbrace{\mathcal{N}(0,\sigma_{\omega}^2)}_{\text{\small noise}},
\end{align}
where $\theta^* = \{m^*, \text{diag}[J^*]\}$ represent unknown parameters, and $\dot{x}$ represents the noisy linear and angular accelerations ($\dot{v}, \dot{\Omega}$). The dynamics derived from first principles (\ref{eq:lin_acc}-\ref{eq:ang_acc}) is incorporated along with Gaussian white noise, $\sigma_{\omega}$, in the surrogate GP modeling. We choose $N = 30$, $\tau_e = 0.05$, along with measurement noise where $\lVert \sigma_x \rVert \leq 3.2 \times10^{-3}$.

\begin{figure}[!b]
\centering
\includegraphics[width=0.85\linewidth]{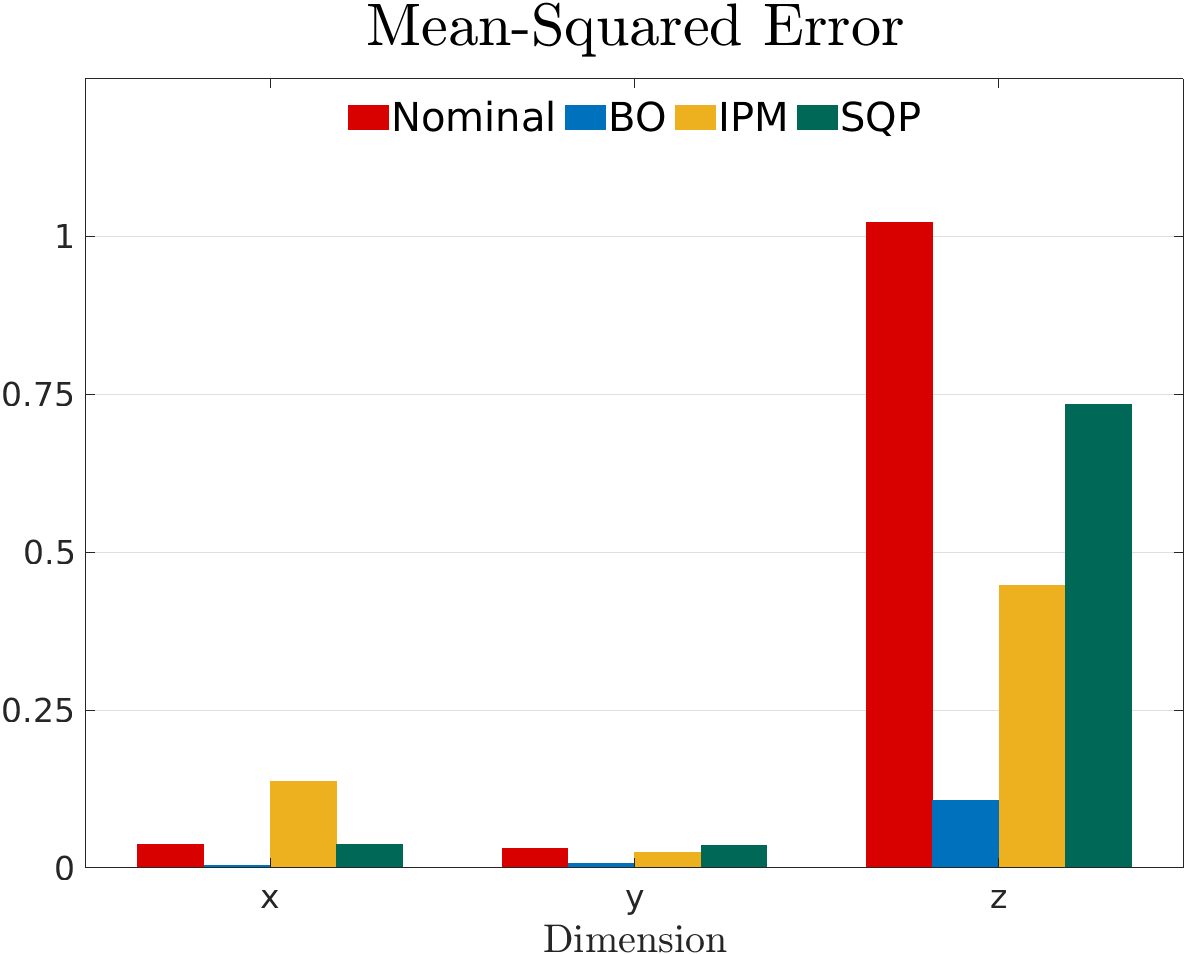}
\caption{Left-to-right: \textbf{Nominal, BO, IPM, SQP}. MSE is shown for each case. Reconfiguring the controller with parameters estimated using BO incurs the least MSE in terms of tracking along every dimension.}
\label{fig:quadrotor_mse}
\end{figure}

\begin{figure}[!b]
\centering
\includegraphics[width=0.85\linewidth]{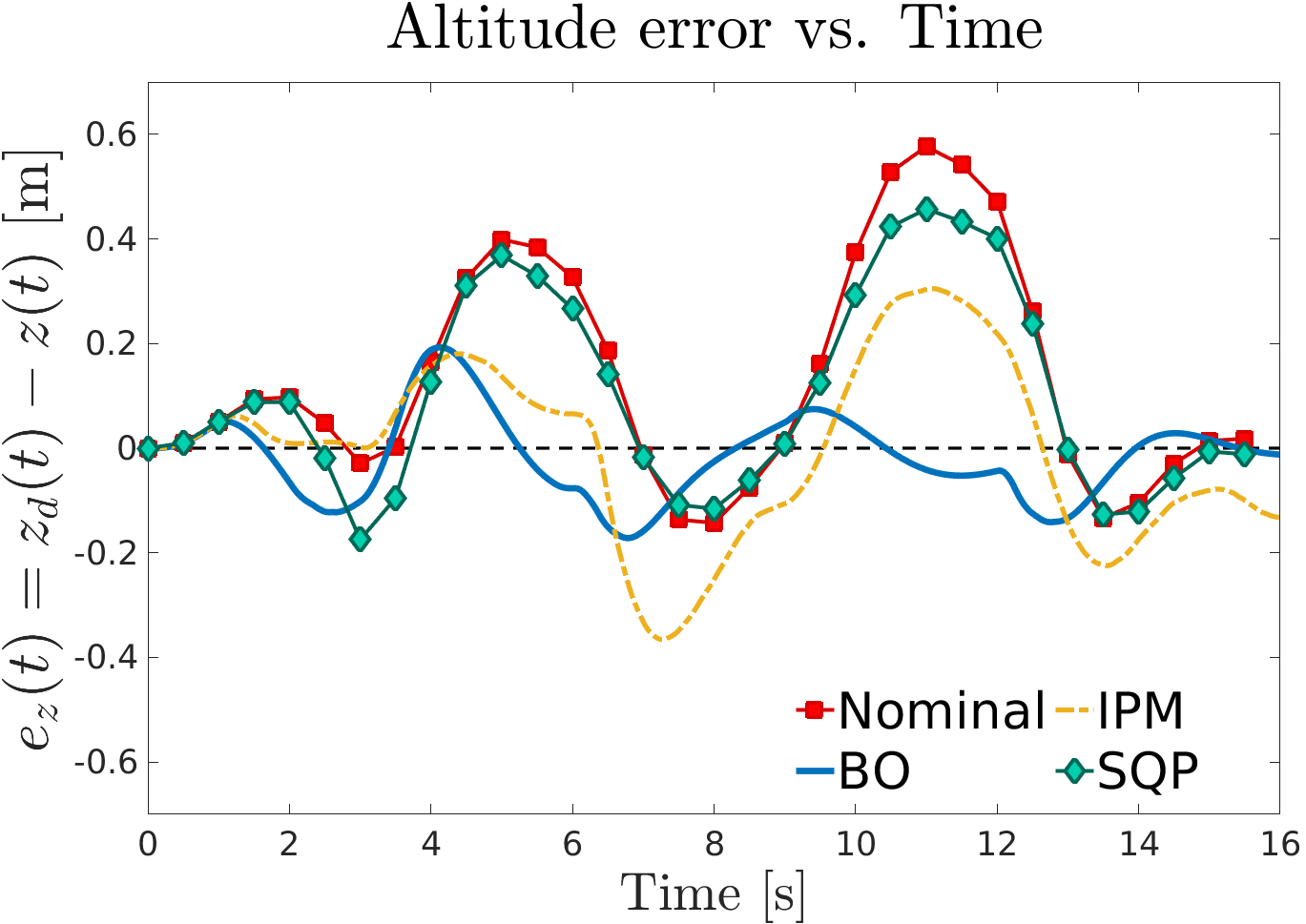}
\caption{The tracking error performance in the altitude domain. The quadrotor exhibits very poor tracking in the nominal case as expected. Our methodology demonstrates the least tracking error due to better parameter estimation.}
\label{fig:quadrotor_tracking}
\end{figure}

\vspace{0.1cm}
\begin{table*}[!h]
\centering
\setlength{\tabcolsep}{1em}
\def\arraystretch{0.65}
\begin{tabular}{ccccccccccccc} \toprule
{Time} & {Optimization} 	& \multicolumn{4}{c}{True values} 	& \multicolumn{4}{c}{Estimated values}	& {Mass} & {Inertial} & {Prediction} \\
\cmidrule(r){3-6}\cmidrule(l){7-10}
 {Instance [\si{\second}]}  &  	{Method}		& 	{m}&{J$_x$}&{J$_y$}&{J$_z$}  & {m}&{J$_x$}&{J$_y$}&{J$_z$} &  {Error} & {Error} & {Time [\si{\second}]}  \\
\midrule
 		& BO 	&  		& 		& 		& 		& 1.58 	& 3.57 	& 3.32 	& 5.53 		& 0.185			& 	\red{1.255}	& 0.324 		\\
1.5 	& IPM 	& 1.39 	& 3.47 	& 3.47 	& 6.77 	& 2.41 	& 1.04 	& 1.03 	& 6.08 		& 1.011			& 	3.508		& 0.028 		\\
		& SQP 	& 		& 		& 		& 		& 1.25 	& 1.10 	& 1.10 	& 2.20 		& \red{0.144}	& 	5.662		& \red{0.015} 	\\
\midrule
 		& BO 	&  		& 		& 		& 		& 1.86 	& 3.91 	& 7.91 	& 10.2 		& \red{0.005}	&	\red{2.966}	& 0.354 		\\
5.0 	& IPM 	& 1.85 	& 6.02 	& 6.02 	& 9.32 	& 1.87 	& 0.95 	& 4.76 	& 2.25 		& 0.024 		&	8.795		& 0.026 		\\
		& SQP 	& 		& 		& 		& 		& 1.25 	& 1.10 	& 1.10 	& 2.20 		& 0.600			&	9.954		& \red{0.013} 	\\
\midrule
 		& BO 	&  		& 		& 		& 		& 1.18 & 2.61 & 5.42 & 9.87 		& \red{0.0004} 	&	3.794		& 0.354 		\\
8.0 	& IPM 	& 1.18	& 3.44 	& 3.44 	& 6.74 	& 1.94 & 1.06 & 3.31 & 4.59 		& 0.7622 		& 	\red{3.212}	& 0.037 		\\
		& SQP 	& 		& 		& 		& 		& 1.25 & 1.10 & 1.10 & 2.20 		& 0.0684 		&	5.621		& \red{0.012}	\\
\midrule
 		& BO 	&  		& 		& 		& 		& 2.12 	& 7.89 	& 5.78 	& 11.50 	& \red{0.018} 	&	\red{2.428}	& 0.354 		\\
11.0 	& IPM 	& 2.10 	& 6.39 	& 6.39 	& 9.69 	& 2.62 	& 1.02 	& 3.10 	& 1.87 		& 0.520 		&	10.032		& 0.028 		\\
		& SQP 	& 		& 		& 		& 		& 1.25 	& 1.10 	& 1.10 	& 2.20 		& 0.850 		&	10.581		& \red{0.014}	\\
\midrule
 		& BO 	&  		& 		& 		& 		& 1.25 	& 3.85 	& 1.57 	& 6.93 		& \red{0.004} 	&	\red{1.844}	& 0.384 		\\
14.0 	& IPM 	& 1.25 	& 3.30 	& 3.30 	& 6.60 	& 2.68 	& 2.31 	& 2.45 	& 3.82 		& 0.090			&	5.064		& 0.028 		\\
		& SQP 	& 		& 		& 		& 		& 1.20 	& 1.67 	& 2.26 	& 2.64 		& 0.087			&	5.312		& \red{0.013}	\\
\bottomrule
\end{tabular}
\vspace{-.25cm}
\caption{\textbf{Parameter estimation accuracy expressed as norm error and prediction time}}\label{tab:accuracy_time}
\vspace{-.6cm}
\end{table*}

We first look at trajectory tracking performance. Applying the framework outlined in \ref{subsec:algorithm}, parameters are estimated online using GPs and the controller in (\ref{eq:quadrotor_controller}) is reconfigured iteratively. Estimation performance is compared against benchmark optimizers such as interior-point method (IPM) and sequential-quadratic programming (SQP). As seen in Figure \ref{fig:quadrotor_mse}, MSE for tracking is the least when reconfigured with parameters estimated using our methodology. Low MSE along $x$ and $y$ is due to transient changes in the mass which has a higher pronounced affect along $z$. Our framework is robust to modeling of noise while finding the global optima and this gives a competing edge over other methods. It also drastically improves tracking performance as seen in Figure \ref{fig:quadrotor_tracking}. Due to limited space, we present the performance along $z$-dimension only since it incurs the highest MSE. This further validates our approach's efficacy in safety-critical applications with improved control performance.

Next, we look at the parameter estimation accuracy using BO and other solvers as tabulated in Table \ref{tab:accuracy_time}. Each solver had domain bounds of $[0.5 \ 4.0]$ for mass and $[0.1 \ 12.0]$ for each inertial axis. IPM and SQP use the nominal parameters as their initial starting points. We look at five time instances, one instance from each time interval in (\ref{eq:time-varying_mass}), for evaluating the estimation accuracy, given as norm error, and prediction time. In all instances, parameters estimated using BO have considerably lesser norm error than other solvers. Even in the case when other solvers have a lower prediction error, BO has marginally close prediction errors ($t=1.5$, $t=8.0$). An important distinction of our approach over other solvers is robustness to noise and estimating parameters for step changes. IPM and SQP at times get stuck at local minima and is unable to estimate these transient changes. SQP is the fastest as expected due to its quasi-newton Hessian approximations while BO has the highest prediction time due to training of GPs and hyperparameter estimation.

\section{Conclusion}\label{sec:conclusion}
Summarizing our work, we have proposed a novel framework using high probabilistic guarantees from GPs to estimate parameters online for safety-critical systems. Using a data-driven approach, BO allows for an efficient search strategy over a response surface to find the global optima. Leveraging the GP posterior mean and variance, the optimizer navigates towards the global optima with an exploration-exploitation mindset. Our formulation does not require gradient approximations and has the competitive edge of incorporating noisy observations in its pursuit to finding the global optima. Finally, we study two cases:- an actuated planar pendulum, investigating a two-dimensional parametric setting highlighting the main aspects of our work, followed by a safety-critical quadrotor system. For the quadrotor, we estimate parameters in a four-dimensional setting with slowly-varying and transient changes to the mass and inertia. We reconfigure the quadrotor controller online with estimated parameters and compare with other benchmark solvers using interior-point methods and hessian approximations. Due to better parameter estimation, the tracking MSE was the least using BO versus other solvers.

Although, our approach has many advantages, we would like to highlight some closing remarks. BO is limited to low-dimensions with increasing estimation time in higher dimensions. This is an active area of research to enable faster tractability using BO. One could leverage techniques such as adaptive mesh search for BO in higher dimensions to query samples quicker \cite{bads_acerbi2017}. In this work, we focused on presenting the main idea for online parameter estimation using GPs. 

\bibliographystyle{ieeetr}
\bibliography{arxiv_lcss20}

\end{document}